\def\year{2022}\relax
\documentclass[letterpaper]{article} 
\usepackage{aaai22}  
\usepackage{times}  
\usepackage{helvet}  
\usepackage{courier}  
\usepackage[hyphens]{url}  
\usepackage{graphicx} 
\usepackage{natbib}  
\usepackage{caption} 
\DeclareCaptionStyle{ruled}{labelfont=normalfont,labelsep=colon,strut=off} 
\usepackage{algorithm}
\usepackage{algorithmic}

\usepackage{newfloat}
\usepackage{listings}
\floatstyle{ruled}
\newfloat{listing}{tb}{lst}{}
\floatname{listing}{Listing}
\nocopyright
\title{Zero-Shot Semantic Segmentation\\ via Spatial and Multi-Scale Aware Visual Class Embedding}

\usepackage{amssymb, booktabs, multirow}
\def\eg{\emph{e.g, }}

\newif\ifaaaifinal
\aaaifinalfalse
\def\aaaifinalcopy{\global\aaaifinaltrue}

\aaaifinalcopy              
\def\aaaiPaperID{451}       

\ifaaaifinal
    \author{
        Sungguk Cha\equalcontrib \textsuperscript{\rm 1}, 
        Yooseung Wang\equalcontrib \textsuperscript{\rm 2}
    }
\else
    \author{
        Anonymous AAAI submission
    }
\fi
\ifaaaifinal
    \affiliations{
        \textsuperscript{\rm 1} MINDsLab Inc., Korea\\
        \textsuperscript{\rm 2} Agency for Defense Development, Daejeon, Korea\\
        sungguk@mindslab.ai, yswang@add.re.kr
    }
\else
    \affiliations{
        Paper ID \aaaiPaperID
    }
\fi

\begin{document}

\maketitle

\begin{abstract}
Fully supervised semantic segmentation technologies bring a paradigm shift in scene understanding. 
However, the burden of expensive labeling cost remains as a challenge. 
To solve the cost problem, recent studies proposed language model based zero-shot semantic segmentation (L-ZSSS) approaches.
In this paper, we address L-ZSSS has a limitation in generalization which is a virtue of zero-shot learning.
Tackling the limitation, we propose a language-model-free zero-shot semantic segmentation framework, Spatial and Multi-scale aware Visual Class Embedding Network (SM-VCENet).
Furthermore, leveraging vision-oriented class embedding SM-VCENet enriches visual information of the class embedding by multi-scale attention and spatial attention.
We also propose a novel benchmark (PASCAL2COCO) for zero-shot semantic segmentation, which provides generalization evaluation by domain adaptation and contains visually challenging samples.
In experiments, our SM-VCENet outperforms zero-shot semantic segmentation state-of-the-art by a relative margin in PASCAL-$5^i$ benchmark and shows  generalization-robustness in PASCAL2COCO benchmark.
\end{abstract}
\section{Introduction}
\begin{figure}[h!]
\centering
\includegraphics[width=1.0\linewidth]{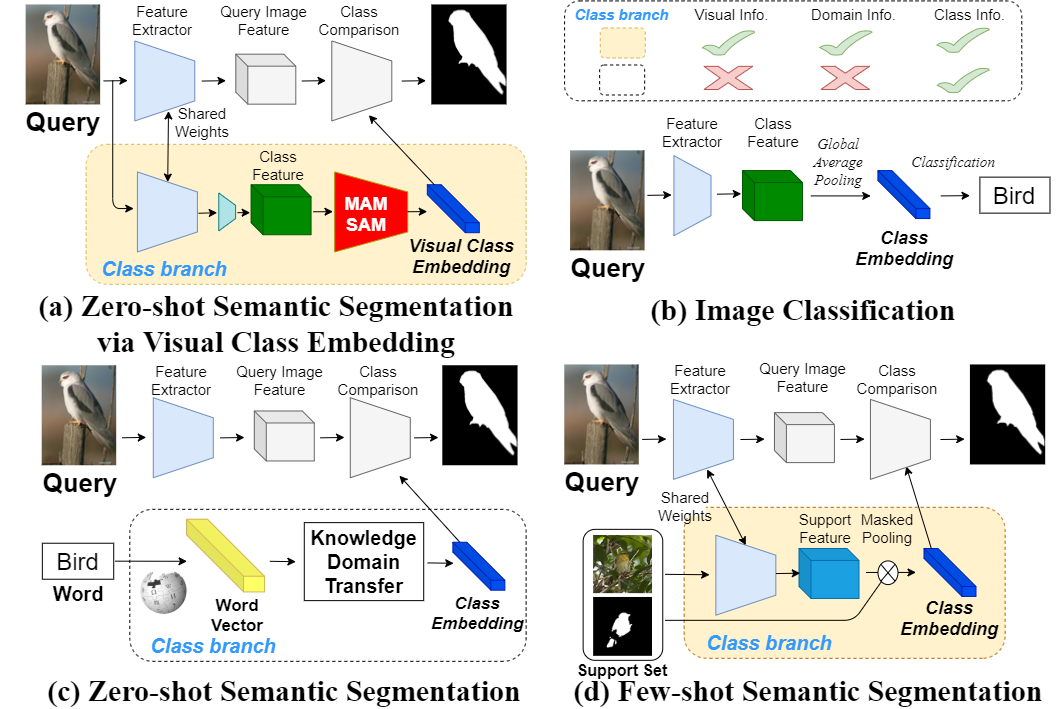}
\caption{
We present a novel language-model-free ZSSS framework, SM-VCENet (a).
SM-VCENet generates class embedding by transferring ImageNet pretrained knowledge onto a query image, resulting in query-domain-aware as in a few-shot semantic segmentation framework.
Our multi-scale attention module and spatial-attention module enrich visual information of the class embedding.
(b) summarizes a general approach to image classification.
(c) and (d) represent L-ZSSS and FSSS approaches.
}
\label{fig:approach}
\end{figure}
By the advent of convolutional neural networks, semantic segmentation methodologies have achieved improvements in accuracy~\cite{deeplabv3,deeplabv3plus,panopticdeeplab,resnest,pspnet} and efficiency~\cite{hardnet,swiftrnnet, frrn, bisenetv2,shelfnet} on the benchmarks such as MS COCO~\cite{coco} and PASCAL VOC~\cite{pascal}.
However, they are impractical in the real-world because they require expensive costs for the annotation of large-scale dataset and fail to predict novel classes that are unseen during the training phase.

To address this problem, \textbf{Z}ero-\textbf{S}hot and \textbf{F}ew-\textbf{S}hot \textbf{S}emantic \textbf{S}egmentation (\textbf{ZSSS} and \textbf{FSSS}, respectively) approaches have arisen to recognize unseen classes with cheaper labeling cost.
ZSSS and FSSS materialize a class representative embedding of an unseen class (hereinafter, \textit{class embedding}) with zero or a few number of target class image(s).
They recognize the novel class by comparing the class embedding and query image feature (shown in Figure~\ref{fig:approach} (c) and (d)).
The recent language model based ZSSS (L-ZSSS) approaches~\cite{z3net, zsvm,spnet} generate a class embedding with a word vector from a language model (the class branch in Figure~\ref{fig:approach} (c)).

We tackle two limitations of the class embedding of L-ZSSS: 
    it cannot cope with domain shift~\cite{torralba2011unbiased} and
    lacks of visual information such as scale and spatial information.
The prior works concentrated on fitting the language based class embedding (LCE) to the training domain, which fails to generalize to an unseen domain.
On the other hand,
    the class embedding of FSSS (Figure~\ref{fig:approach} (d))
        shares the same domain as the query image 
        and contains visual information.
Our experiments reveal domain shift causes changes in visual class embedding (Figure~\ref{figure:affinity}), thus L-ZSSS fails to generalize to unseen domain.
In our experiments, L-ZSSS showed poor performances in visually challenging cases (\eg multiple, multi-scale objects and complex object in noisy background).

In this paper, overcoming the limitations, we propose \textbf{S}patial \& \textbf{M}ulti-scale aware \textbf{V}isual \textbf{C}lass \textbf{E}mbedding \textbf{Net}work (\textbf{SM-VCENet}).
SM-VCENet generates class embedding by transferring vision-oriented knowledge onto a query image.
As the query image derives the class embedding, it follows the query domain distribution, becoming \textit{query-domain-aware}.
Our spatial attention module (SAM) and multi-scale attention module (MAM) enhance the class embedding to have rich visual information.
In PASCAL-$5^i$ test benchmark, SM-VCENet achieves 46.5\% mIoU score, outperforming ZSSS state-of-the-art (SOTA) SPNet~\cite{spnet_pascal} (44.9\% mIoU) by a great margin.

Moreover, we present a novel challenging ZSSS benchmark PASCAL2COCO that evaluates generalization ability and contains visually challenging samples.
Although generalization ability is an important virtue of zero-shot learning, existing benchmark PASCAL-$5^i$~\cite{pascal} cannot handle it.
Proposed PASCAL2COCO benchmark evaluates generalization ability by conducting a domain adaptation task (train on PASCAL-$5^i$ train set and test on both PASCAL-$5^i$ test set and COCO-$20^i$~\cite{coco} test set).
Additionally, it provides visually challenging samples such as multiple objects, multi-scale objects and complex objects in noisy background. 

Our contributions are summarized as follows:
\begin{enumerate}
\item 
We propose a novel language-model-free ZSSS framework SM-VCENet, addressing domain-agnostic class embedding problem of L-ZSSS.
By transferring a vision-oriented knowledge to a query image, SM-VCENet achieves query-domain-aware class embedding.
In our experiments, SM-VCENet shows \textit{generalization-robustness} in domain adaptation problem.
\item
We propose SAM and MAM that enrich visual information of the class embedding.
We show that SAM and MAM achieve significant improvements in \textit{visually challenging} cases over ZSSS SOTA by a great margin.
\item
We first present ZSSS benchmark PASCAL2COCO that evaluates generalization ability and contains visually challenging samples.
In a sense that generalization is the most important idea of zero-shot learning, our proposed benchmark provides valid generalization measurement.
\end{enumerate}
    \begin{figure*}[ht]
\centering
\includegraphics[width=1.0\linewidth]{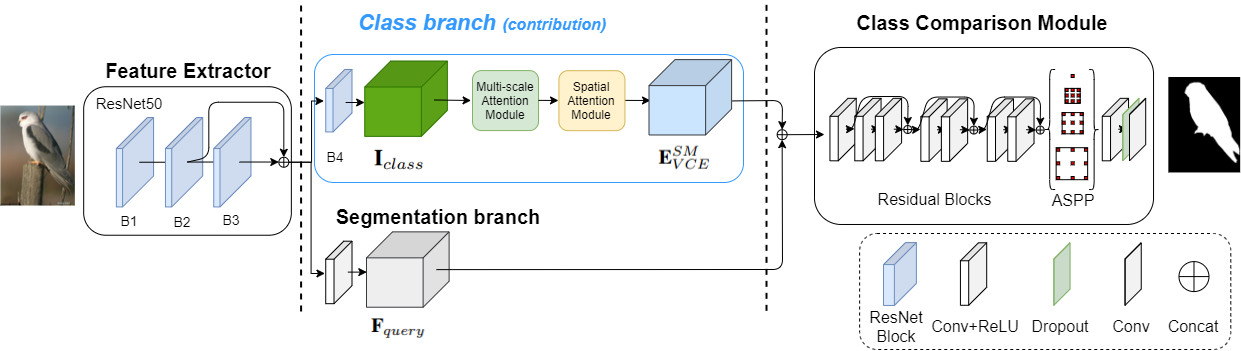}
\caption{
An overview of our proposed language-model-free ZSSS framework, SM-VCENet.
}
\label{fig:architecture}
\end{figure*}
\section{Related Works}
\textbf{Semantic Segmentation}
Semantic segmentation requires multi-scale and spatial information understanding. 
In order to grasp multi-scale information, multi-scaled feature extraction and aggregation approaches~\cite{sppnet, sift, pspnet} have been researched.
Scale-invariant feature transform (SIFT) method~\cite{sift} extracts a feature from multi-scaled image via Difference of Gaussian.
Motivated, SPPNet\cite{sppnet} and PSPNet~\cite{pspnet} utilized multi-scale feature extraction for modern deep convolutional neural networks (DCNNs).
Spatial information understanding, including localization~\cite{fcn, hrnet} and global context information~\cite{nnn, psanet}, for DCNNs has been proposed.
FCN~\cite{fcn} secured localization features by reducing resolution deductions in DCNNs.
Further, HRNet~\cite{hrnet} dramatically reduced the deduction by high-resolution convolutions.

PSANet~\cite{psanet} contained point-wise spatial attention to relax the local neighborhood constraint.
Recently, the non-local operation based networks~\cite{nnn, cgnl, apnl} have been proposed to grap global context of features by enlarging the receptive field. 
Non-local operation~\cite{nnn}  captures long-range dependencies by computing a output pixel value from weighted summation of all input feature values.
Moreover, asymmetric~\cite{apnl} and compact-generalized~\cite{cgnl} non-local network decreased complexity computation for the matrix multiplication by proposing asymmetric matrix dimension and compact representation for multiple kernels. 
In this work, we adopted multi-scale feature extraction and non-local block~\cite{nnn} for \textit{multi-scale} and \textit{spatial} information aware class embedding. 

\textbf{Zero-(Few-)Shot Semantic Segmentation}
Shown in Figure~\ref{fig:approach}, ZSSS and FSSS approaches share the same framework with two major components:
(1) class embedding generation (the class branch in Figure~\ref{fig:approach}),
and (2) class comparison between segmentation feature and class embedding.

First, FSSS and ZSSS approaches worked on generating expressive class embedding.
Making class embedding to contain meaningful visual information in FSSS,
PPNet~\cite{ppnet} extracts part-aware class embedding grouped by super pixel by SLIC~\cite{slic}. 
PGNet~\cite{pgnet} utilized attention masked pooling.
Recently, ASR~\cite{asr} proposed anti-aliasing representations between categories and SAGNN~\cite{sagnn} presented scale aware graph neural network to encode class embedding. 
On the other hand, ZSSS that has no supporting visual source generates class embedding with language context knowledge, word vector. 
ZS3Net~\cite{z3net} learns to generate synthetic features with word2vec and visual feature. 
ZSVM~\cite{zsvm} maps the word vector into visual semantic space by variational sampling.
GZS3~\cite{gzs3} proposed jointly training word vector and vision model.

Second,
comparing the class embedding and the segmentation feature is actively studied.
In FSSS, 
AMP~\cite{amp} showed adaptively weighted classifier with the label masked pooling on support images.
CRNet~\cite{liu2020crnet} and PANet~\cite{panet} added query-support symmetric branch methods that mutually learn from both query image and support image.
CANet~\cite{canet} presented the pixel-wise comparison framework with iterative optimization.
Feature weight method~\cite{featweight} guided ensemble inference in the multi-shot setting. 
On the other hand, ZSVM and ZSI~\cite{zsvm, zsis} conducted ZSSS as in FSSS by considering word vector as class embedding. 
While, SPNet~\cite{spnet} conducted ZSSS by directly using word vector as classifier weights.
In this work, 
we introduce a novel approach to generate class embedding even without side information such as the word vector in the ZSSS setting.

\begin{figure}[ht]
\centering
\includegraphics[width=1.0\linewidth]{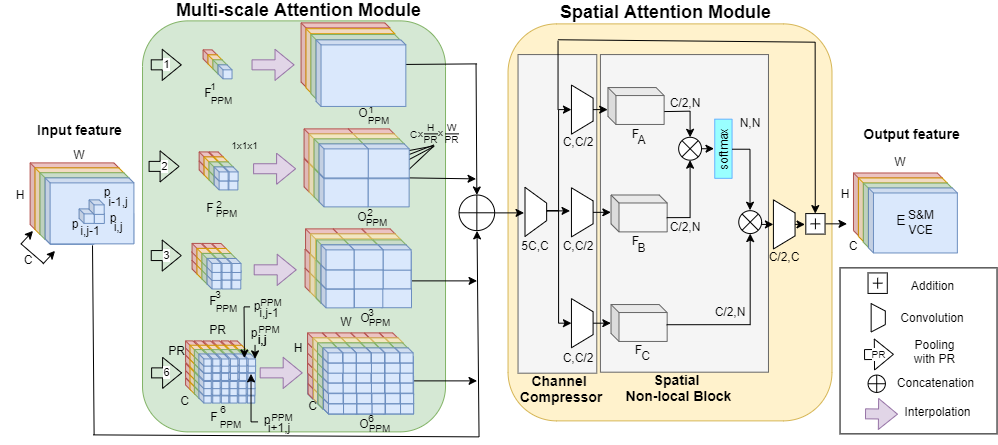}
\caption{
Multi-scale attention module (MAM) and spatial attention module (SAM) overview. 
MAM produces multi-scale aware feature map by pyramidal pooling.
SAM compresses the concatenated feature and gives global context information via a non-local module.
}
\label{fig:sma}
\end{figure}
\section{Method}

\subsection{Visual Class Embedding (VCE) Motivation}
For predicting an unseen class, we need a class representative expression (class embedding) as we cannot have a classifier that solves the regression problem that if the query feature belongs to the unseen class.
The class embedding can be in the form of constant vector, as a word is expressed in the form a vector from the word vector language model~\cite{glove}.
However the prior works~\cite{z3net, zsvm, spnet} assumed the lingual knowledge can be directly used in the visual task, we think the two knowledge must form different distribution.
Thus, instead, we propose to use ImageNet~\cite{imagenet} pre-trained knowledge for the class embedding. 
Shown in the Figure~\ref{fig:approach} (b), the well known image classification task is based on the class embedding generation by ResNet~\cite{resnet} and VGGNet~\cite{vgg}.
Our proposed Visual Class Embedding (VCE) is from the query image with ImageNet pretrained networks. 
By freezing the pretrained backbone network, we preserve the visual domain knowledge.
While the class embedding from supporting images of FSSS are visual that includes visual knowledge (\eg topological, spatial and scale information), the class embedding from the L-ZSSS approaches cannot contain such visual knowledge as a word vector cannot imply them.
At the best of our knowledge, we are the first proposing visual knowledge oriented class embedding for zero-shot learning.
Maximizing the advantages of visual knowledge, we propose MAM and SAM for visual class embedding. 

\subsection{Overall Architecture of SM-VCENet}
SM-VCENet consists of three main parts: a feature extractor, class and segmentation branches, and class comparison module (CCM). 
Figure \ref{fig:architecture} illustrates the overall architecture of SM-VCENet. 
We use ImageNet-pretrained ResNet50 to extract the query image feature $F_{query}$.
The class branch generates VCE.
CCM conducts semantic segmentation by comparing $F_{query}$ and VCE in pixel level with three residual blocks and ASPP~\cite{deeplabv3}.

We extract the query image feature by ImageNet pretrained ResNet50~\cite{resnet} without updating the parameters.
Given a set of ResNet blocks $\{\textbf{B}_{1},\textbf{B}_{2},\textbf{B}_{3}$ and $\textbf{B}_{4}\}$, the corresponding features $\textbf{F}_{1}, \textbf{F}_{2}$ and $\textbf{F}_{3}$ are extracted from the RGB query image $\textbf{X} \in \mathbf{R}^{3 \times H_{input} \times W_{input}}$ as follows: $ \textbf{F}_{i} = \textbf{B}_{i}(\textbf{F}_{i-1})$ where $i \in {2,3,4}$ and $ \textbf{F}_{1} = B_{1}(X) $ where $H_{input}$, and $W_{input}$ are the input image height, and width. 
Inspired by the previous study~\cite{canet} of an output feature from ResNet~\cite{resnet} backbone network for FSSS, we compute the query features $\textbf{F}_{query}$ for the segmentation branch by concatenating features $\textbf{F}_{2}$ and $\textbf{F}_{3}$, and the input of class branch from $\textbf{B}_{4}$: 
\begin{equation} 
\textbf{I}_{class} = \textbf{B}_{4}(\textbf{F}_{2}\oplus \textbf{F}_{3})
\end{equation} 
\begin{equation} 
\textbf{F}_{query} = f(\textbf{F}_{2}\oplus \textbf{F}_{3})
\end{equation} where $f$ is the $3 \times 3$ convolution operation. 

\textbf{Class branch}
The class branch creates class embedding that contains both spatial information and multiple size of compressed features for predicting multi-scale objects throughout two modules: multi-scale attention module (MAM) and spatial attention module (SAM).
Figure \ref{fig:sma} represents the overall operations for MAM and SAM. 

MAM extracts implicit information of features of various sizes through pooling with multiple ratios.
Inspired by PSPNet~\cite{pspnet}, we adopt the early stage of pyramid spatial pooling module to compact multi-scaled information. 
Given an input $ \textbf{I}_{class} \in \mathbb{R}^{C \times H \times W} $, let $p_{c,i,j} \in \textbf{I}_{class}$  the each pixel value where the c, i, j represent the position of it in the three-dimensional input feature. 
MA compresses it by the pooling with pooling ratio of (1, 2, 3, 6). Each feature maps pooled with $PR$ pooling ratio divide the $H\times W$ size feature map into the $PR^2$ number of regions and extract one representative values for each region. Each compressed value is calculated by averaging pixel values in the specific region: 
\begin{equation} 
p_{c,i,j}^{PPM} = \frac{ \sum_{h=1}^{i+i*(H/NR-1)}\sum_{w=1}^{j+j*(W/NR-1)} p_{c,i,j}}{NR}
\label{eq:5} 
\end{equation} where number of regions $NR = \frac{H \times W}{PR^2} $ and $0 < i,j < PR$.
Then we have $PR^2$ sized of two-dimensional output for each pooling ratio. The output feature map with PR pooling ratio is calculated as follows: 

\begin{equation} 
\textbf{F}_{PPM}^{PR}=\sum_{c=1}^{C} \sum_{w=1}^{PR}\sum_{h=1}^{PR}p_{c,h,w}
\end{equation} 
We have four different size of outputs $\textbf{F}_{PPM}^{1}, \textbf{F}_{PPM}^{2}, \textbf{F}_{PPM}^{3}, \mbox{~and~} \textbf{F}_{PPM}^{6}$ with the pooling ratios in (1, 2, 3, 6). 
Note that each $p_{c,i,j}$ in $\textbf{F}_{PPM}^{PR}$ represents the distinct compressed information of input feature map which helps the model to predict multi-scaled objects.
Therefore, MAM considers totally $1 \times 1 + 2 \times 2 + 3 \times 3 + 6 \times 6 = 50$ number of compressed information from multi-scaled features.  
We expand all the outputs, $\textbf{F}_{PPM}^{i} \in \mathbb{R}^{C \times PR \times PR}, i \in {1,2,3,6}$ to the $\textbf{O}_{PPM}^{i} \in \mathbb{R}^{C \times H \times W}$ where $ i \in {1,2,3,6}$.
We concatenate them with the input feature $\textbf{I}_{class}$.
Therefore, we have the final output of MAM as follows:
\begin{equation} 
\textbf{O}_{M} = \mbox{Cat}(\textbf{O}_{PPM}^{1}, \textbf{O}_{PPM}^{2}, \textbf{O}_{PPM}^{3}, \textbf{O}_{PPM}^{6}, \textbf{I}_{class}) 
\end{equation} 

SAM uses
the MA output that combines feature information of different sizes to interlink dependencies of each multi-scaled information.
SAM calculates related information of those having spatial information of different sizes through non-local~\cite{nnn} operations.
Compressor in SAM first compressed $\textbf{O}_{M} \in \mathbb{R}^{5C \times H \times W}$ to the dimension of $C/2 \times H \times W$ for the effective non-local operations by two stages of convolution.
For the first stage, we decrease the channel of feature from 5C to C.
We separate the output of the first stage into three branches and compute convolutions for each features with distinct weight matrices. 
To do the matrix multiplication between the three-dimensional features in order to connect dense relationships, we first view the dimension of each input from $C \times H \times W$ into $C \times N$ where $ N = H \times W$. 
We multiply the feature $\textbf{F}_{A} \in \mathbb{R}^{C \times N}$ and transposed feature $\textbf{F}_{B} \in \mathbb{R}^{N \times C} $ to output $\textbf{F}_{D}$.
The second multiplication output $\textbf{F}_{M}$ is computed by the softmax output of $\textbf{F}_{D}$ and $\textbf{F}_{C}$.
Note that all features in the non-local operation ($\textbf{F}_{A}$, $\textbf{F}_{B}$,  $\textbf{F}_{C}$, $\textbf{F}_{D}$, and output $\textbf{F}_{M}$) represent the condensed multi-scale information of feature. 
Therefore, two multiplications of each feature strengthen connections for global context between multi-scale information.
We recover the reduced channel of $\textbf{F}_{M}$ from $C / 2 $ to the C. The final output of Spatial non-local block is the concatenation of $\textbf{O}_{M}$ and $\textbf{F}_{M}$ as follows: $ \textbf{E}_{VCE}^{SM} = Concat(\textbf{F}_{M},f( \textbf{O}_{M}))$

\textbf{Class comparison module (CCM)}
CCM performs semantic segmentation by comparing the class embedding ${E}_{VCE}^{SM}$ and the query feature $F_{query}$, acting like the segmentation decoder in \cite{deeplabv3}.
Given the two features, it concatenates all the vectors in pixel-wise, which preserves the mutual location information.
For efficient implementation, we reduce the number of the channel of the concatenated feature from $256 + C$ to $256$ with a $1 \times 1$ convolution layer.
Next, CCM solves a regression problem that if a feature vector of a pixel of $F_{query}$ belongs to the same class.
With the following three sequential basic residual blocks~\cite{resnet} where a residual block consists of two $3 \times 3$ convolutional layers with skip connection, CCM compares the two features.
Finally, after ASPP~\cite{deeplabv3}, CCM results a predicted binary mask $M \in \{0, 1\}^{H_{input} \times W_{input}}$.

Note that each convolutional layers in CCM are followed by ReLU activation function without batch normalization~\cite{bn}.
However, in zero-(few-)shot learning setting where each sample of a mini-batch may have a different target class, the layers from different batches may have different distribution expressing their own target class.
Thus, normalizing a layer with respect to batch smooths the expressiveness of the layer about the target class.

\begin{figure}
    \centering
    \includegraphics[width=0.475\textwidth]{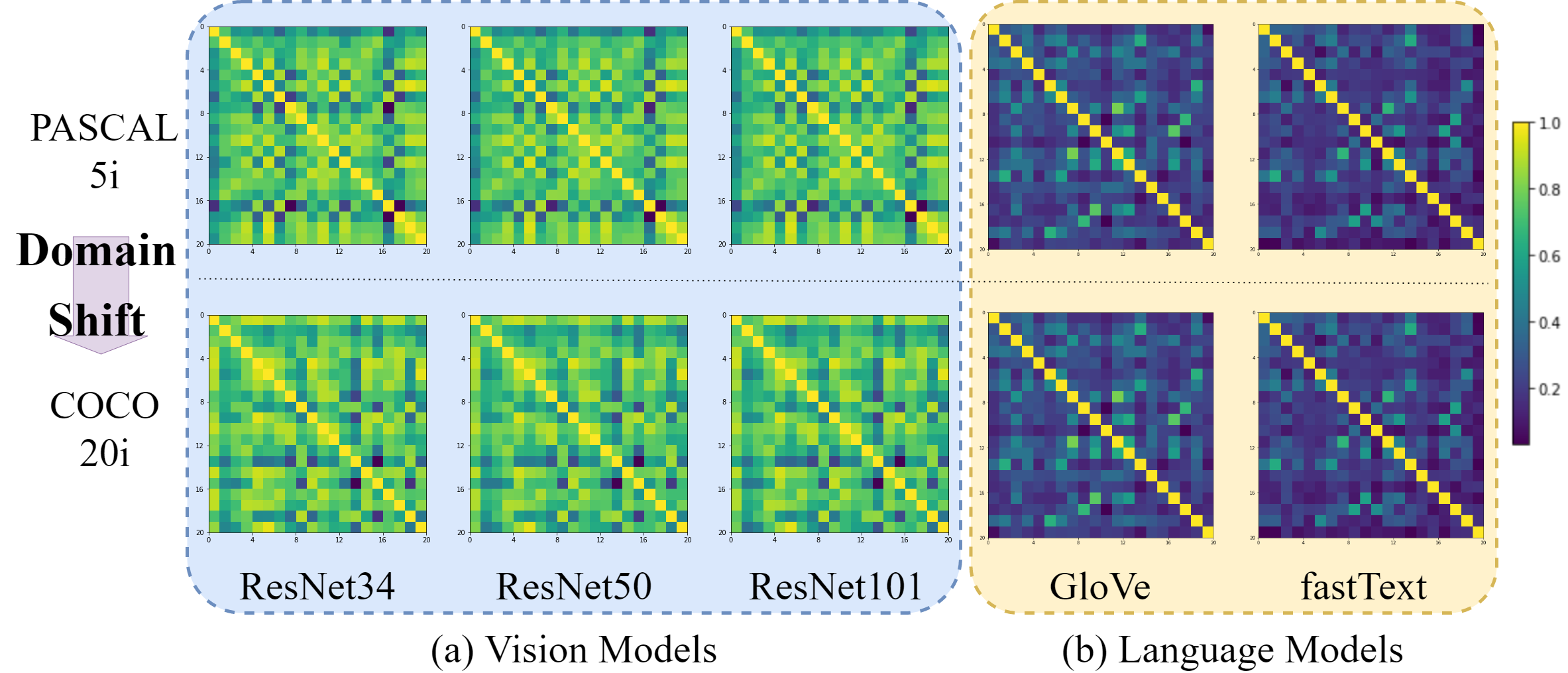} {}
    \caption{
        Class embedding similarity matrices over 20 PASCAL categories.
        The categories are assigned top-to-bottom and left-to-right.
    }
    \label{figure:affinity}
\end{figure}

\begin{table*}[ht]
\small
\centering
\setlength{\tabcolsep}{4pt}
\begin{tabular}{@{}cc|cc|c|c|cccc|c|c@{}}
\toprule
\multirow{2}{*}{Models} & \multirow{2}{*}{Trained} & \multicolumn{2}{c|}{External Knowledge} & \multicolumn{2}{c|}{PASCAL-$5^i$} & \multicolumn{6}{c}{COCO-$20^i$} \\
\cmidrule{3-12}
& & ImageNet & Word & mIoU & Average & $i=0$ & $1$ & $2$ & $3$ & mIoU & Average \\
\midrule
\multirow{4}{*}{DeepLab}
& PASCAL-$5^0$ & \checkmark &  & 36.49 & \multirow{4}{*}{38.59} & 33.32 & 30.3  & 25.30  & 31.85  & 30.20 & \multirow{4}{*}{30.53} \\
& PASCAL-$5^1$ & \checkmark &  & 47.85 & & 34.81 & 32.51 & 25.92 & 32.16 & 31.35 & \\
& PASCAL-$5^2$ & \checkmark &  & 39.75 & & 33.64 & 34.29 & 25.39 & 33.02 & 31.59 & \\
& PASCAL-$5^3$ & \checkmark &  & 30.28 & & 31.88 & 31.74 & 22.68 & 29.56 & 28.97 & \\
\midrule
\multirow{4}{*}{SPNet}
& PASCAL-$5^0$ & \checkmark  &\checkmark  & 43.72 &    &  26.61 & 23.14 & 22.92 & 24.79 & 24.37   &    \\
& PASCAL-$5^1$ & \checkmark &\checkmark  & 60.42 & 45.06 & 30.13 & 29.55 & 23.45   & 22.23 & 26.34   & 24.51 \\
& PASCAL-$5^2$ & \checkmark &\checkmark  & 40.44 & (+6.47) & 21.12 & 24.43 & 25.64 & 23.33 & 23.63   & (-6.02) \\
& PASCAL-$5^3$ & \checkmark &\checkmark  & 35.64 &  & 26.57 & 22.07   & 20.99 & 25.23 & 23.72   &  \\
\midrule
\multirow{4}{*}{ZSVM}
& PASCAL-$5^0$ & \checkmark &\checkmark  & 47.07 &    & 40.94 & 36.08 & 30.44 & 32.09 & 34.89   &    \\
& PASCAL-$5^1$ & \checkmark &\checkmark  & 59.73 & 45.32 & 34.54 & 31.88 & 26.82   & 27.76 & 30.25   & 31.65 \\
& PASCAL-$5^2$ & \checkmark &\checkmark  & 36.54 & (+6.73) & 37.00 & 37.90 & 33.07 & 30.64 & 34.65   & (+1.12) \\
& PASCAL-$5^3$ & \checkmark &\checkmark  & 37.92 &  & 35.34 & 23.72 & 22.69 & 25.56 & 26.83  &  \\
\midrule
& PASCAL-$5^0$ & \checkmark &  & 48.09  &  & 37.60  & 39.25  & 39.92  & 41.81 & 39.65 &  \\
SM-VCENet & PASCAL-$5^1$ & \checkmark & & 54.21 & \textbf{46.51} & 42.69  & 45.69  & 40.44  & 42.77 & 42.89 & \textbf{41.34} \\
(Ours) & PASCAL-$5^2$ & \checkmark &  & 43.12 & (\textbf{+7.92}) & 39.78  & 41.08  & 38.16  & 45.59  & 41.15 & \textbf{(+10.81)} \\
& PASCAL-$5^3$ & \checkmark & & 40.62 &  & 39.70  & 45.07  & 38.80  & 43.12 & 41.67 &  \\
\bottomrule
\end{tabular}
\caption{
Zero-shot semantic segmentation results for PASCAL2COCO task. 
External knowledge column explains usage of ImageNet pretrained weights and word embedding.
All models share the same ImageNet pretrained ResNet50.
}
\label{table:main}
\end{table*}
\begin{table}[ht]
\centering
\resizebox{.49\textwidth}{!}{
\begin{tabular}{cccccccccc}
\toprule
\multirow{2}{*}{Models} & \multirow{2}{*}{Backbone} & \multicolumn{3}{c}{External Knowledge}  & \multicolumn{5}{c}{PASCAL-$5^i$} \\
& & ImageNet & Word & Image & $i=0$  & $1$  & $2$  & $3$ & Mean\\
\midrule
OSVOS & VGG16 & \checkmark & & \checkmark & 24.9 & 38.8 & 36.5 & 30.1 & 32.6 \\
DeepLab & ResNet50 & \checkmark & & & 36.5 & 47.9 & 39.8 & 30.3 & 38.6 \\ 
OSLSM & VGG16 & \checkmark & & \checkmark & 33.6 & 55.3 & 40.9 & 33.5 & 40.8 \\
co-FCN & VGG16 & \checkmark & & \checkmark & 36.7 & 50.6 & 44.9 & 32.4 & 41.1 \\ 
Z3Net & ResNet101 & \checkmark &  \checkmark & &  - & - & - & - & 41.3 \\ 
ZSVM & VGG16 & \checkmark & \checkmark & & 39.6 & 52.6 & 41.0 & 35.6      & 42.2 \\ 
AMP-1 & VGG16 & \checkmark & & \checkmark & 37.4 & 50.9 & 46.5 & 34.8 & 42.4 \\ 
AMP-2 & VGG16 & \checkmark & & \checkmark & 41.9 & 50.2 & \textbf{46.7} & 34.7 & 43.4 \\ 
SPNet & ResNet50 & \checkmark & \checkmark &  & 43.7 & \textbf{60.4} & 40.4 & 35.6 & 45.1 \\ 
ZSVM & ResNet50 & \checkmark & \checkmark &  & 47.1 & 59.7 & 36.5 & 37.9 & 45.3 \\
\midrule
SM-VCENet & ResNet50 & \checkmark & & & \textbf{48.1} & 54.2 & 43.1 & \textbf{40.2} & \textbf{46.5} \\ 
\bottomrule
\end{tabular}
}
\caption{
Zero-shot semantic segmentation results for PASCAL-$5^i$ task.
The Image column marks one-shot approaches that require a supplementary image.
}
\label{table:validation}
\end{table}

\begin{figure*}[ht]
  \centering
    \includegraphics[width=0.95\textwidth]{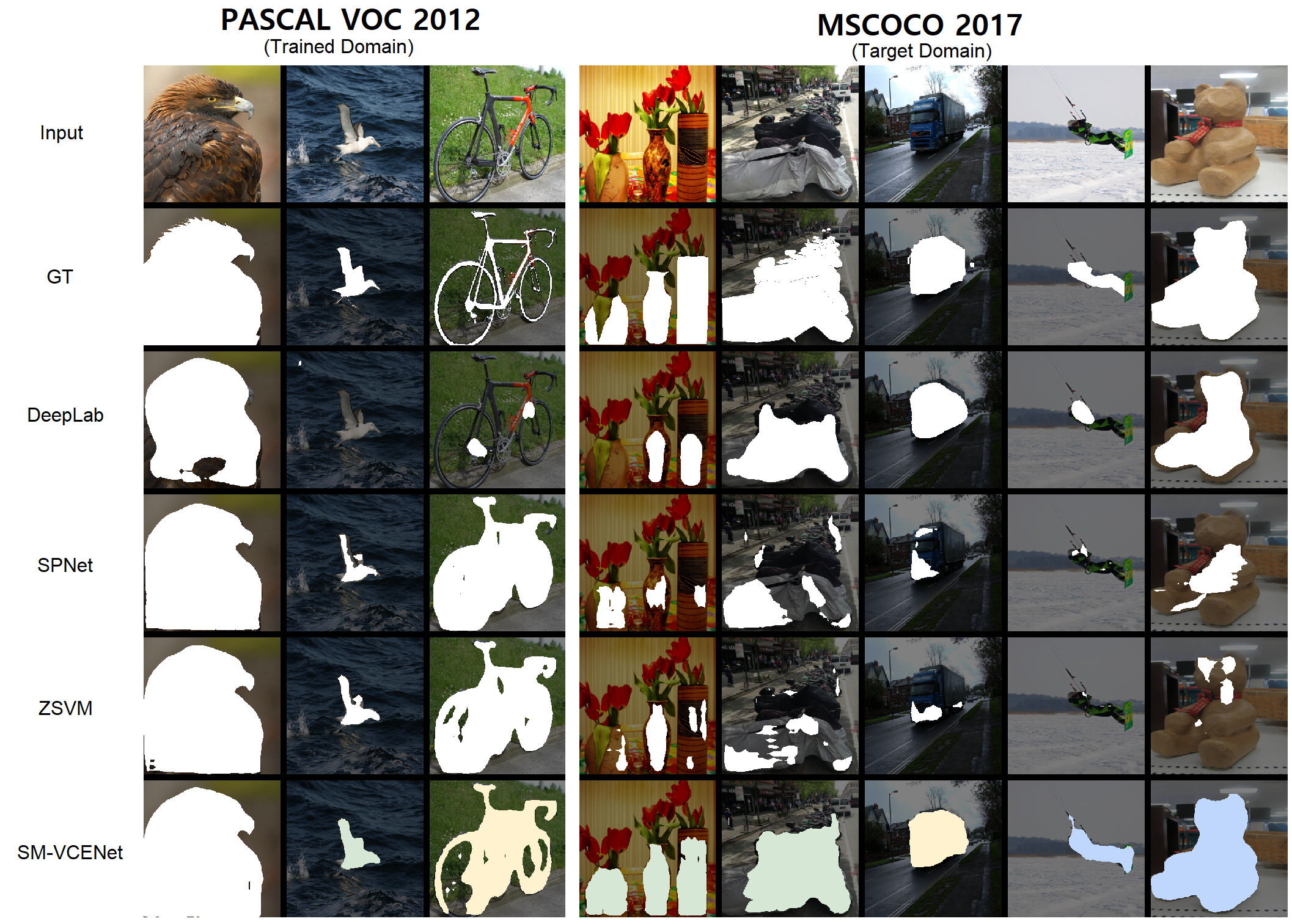} {}
  \caption{
Qualitative comparisons of ZSSS methods on PASCAL VOC 2012 and MS COCO 2017 test images.
GT denotes ground truth.
We highlighted the strengths of our SM-VCENet: multi-scale understanding (light green), spatial understanding (yellow) and generalization (blue).
}
\label{figure:qual}
\end{figure*}

\section{Experiments}
\label{exp}

\subsection{Implementation Details}
We conduct all experiments in PyTorch~\cite{pytorch} framework, following the settings of \cite{zsvm, canet}.
We employ the mean of cross entropy loss over all spatial locations in the output feature map.
We adopt the Stochastic Gradient Descent (SGD) optimizer with mini-batches through 200 epochs on PASCAL-$5^i$ train set.
We set the initial learning rate to 0.0025, momentum as 0.9 and weight decay as 0.0005.

\textbf{Word vector}
We used 300-dimensional word embedding vectors of GloVe \cite{glove} pretrained on Common Crawl with 2.2M words.
Following ZSSS settings~\cite{zsvm, spnet}, for the word embeddings of the classes expressed in multiple words in both PASCAL-5$^i$ and COCO-20$^i$ classes, we averaged the word vectors (word vectors of "potted plant" and "tv/monitor" are the mean vector of each word in the label name) or simplified label name ("dinningtable" as "table").


\subsection{Domain Shift and Class Representation} 
\label{domainshift}
We show class embedding changes when domain is shifted in Figure~\ref{figure:affinity}.
It supports our claim that domain awareness is necessary for ZSSS to generalize to unseen categories.

Figure~\ref{figure:affinity} shows the class representation affinity matrices of vision (ResNets) and language (GloVe and fastText) models in PASCAL-$5^i$ and COCO-$20^i$ datasets.
We use cosine similarity measure to express relationships between classes.
We generate class representations of ImageNet pretrained ResNet34, ResNet50 and ResNet101 from PASCAL-$5^i$ and COCO-$20^i$ datasets.
Given PASCAL-$5^i$ with 20 categories $X = \mathop{\cup}_{i}^{20} \mathop{\cup}_{j \in |C_i|} \{ I_{i, j} \}$ and a model $f: I \rightarrow r$ that encode an image $I$ into a $2048$ dimensional class representation $r$, the i-th category representation $R_i$ is
\begin{equation}
    R_i = \frac{1}{|C_i|}\sum_j^{|C_i|} f(I_{i, j})
\end{equation}
We use two word embedding models (GloVe and fastText).
As shown in the Figure~\ref{figure:affinity}, class representations in CNN change when input image domain shifts. 
In addition, class affinity matrices of vision models show the same pattern of the similarities under the same domain.
On the other hand, the class representations of language models are constant regardless of domain shift.
\subsection{PASCAL-$5^i$ (Trained Domain)}
\label{pascal5i}
The PASCAL-$5^i$~\cite{pascal} dataset is consisted of images from PASCAL VOC 2012~\cite{pascal} and additional annotations from SDS~\cite{sds}.
It has 41,040 training images and 4,000 test images.
It consists of four sub-datasets, dividing the 20 object classes in PASCAL.
Each sub-dataset contains 15 training (seen) classes 5 test (unseen) classes. 
To achieve fairness in comparing with FSS approaches,
we followed the same PASCAL-$5^i$ FSSS setting~\cite{ppnet, amp, panet, pgnet, canet}.

\textbf{Quantitative results}
\label{validate}
Table~\ref{table:validation} compares our SM-VCENet with zero-(one-)shot semantic segmentation approaches.
DeepLab~\cite{deeplabv3} is the backbone of SM-VCENet which is the same architecture in Figure~\ref{fig:architecture} without class branch.
Zero-shot baselines are Z3Net~\cite{z3net}, ZSVM~\cite{zsvm} and SPNet~\cite{spnet}. 
SPNet and ZSVM with ResNet50 in Table~\ref{table:validation} are our re-implementation sharing the same backbone of our SM-VCENet. 
One-shot baselines are OSVOS~\cite{osvos}, OSLSM~\cite{oslsm}, co-FCN~\cite{cofcn} and AMP~\cite{amp}.
Our SM-VCENet records the best accuracy achieving $46.5\%$ mIoU score in ZSSS.

\textbf{Qualitative results}
Figure \ref{figure:qual} qualitatively compares the performances of SM-VCENet with DeepLab, SPNet and ZSVM on PASCAL test set images.
It contains following three PASCAL-representative examples:
large (the first column, a big bird), 
small (the second column, a small bird)
and complex (the third column, a bicycle) cases.
SPNet, ZSVM, and SM-VCENet improve the performance of their backbone (DeepLab). 
Compared with the L-ZSSS, SM-VCENet segments the delicate bicycle wheel better. 

\subsection{COCO-$20^i$ (Target Domain)}
\label{coco20i}
COCO-$20^i$ consists of 12,468 test images from MS COCO 2017~\cite{coco} with 80 classes.
We followed the same COCO-$20-i$ setting as \cite{featweight, panet}.
It has four sub-datasets (COCO-$20^0$,-$20^1$,...), where each dataset has 60 train classes and 20 test classes.

\textbf{Motivation} 
We present a unified ZSSS benchmark (PASCAL2COCO) with two advantages.
First, PASCAL2COCO benchmark contains an evaluation in generalization ability which is an important virtue of zero-shot learning.
PASCAL2COCO evaluates the generalization-robustness of a model by conducting a domain adaptation that trains on PASCAL-$5^i$ and tests on COCO-$20^i$. 
Second, PASCAL2COCO provides elaborate measurements on visually challenging samples. 
COCO-$20^i$ contains cases of multiple objects, multi-scaled objects and complex objects in noisy background.

\textbf{Quantitative results}
\label{coco:quan}
Table~\ref{table:main} compares ZSSS performances of our SM-VCENet with Deeplab, SPNet and ZSVM on PASCAL2COCO.
Note that SM-VCENet, SPNet and ZSVM share the same backbone DeepLab. 
Therefore, we can see generalization ability of the models by observing improvement of performance over DeepLab.
SM-VCENet shows generalization-robustness by steadily improving in both PASCAL-$5^i$ average mIoU score and COCO-$20^i$ domain adaptation average mIoU score with a great margin (+7.92 mIoU in PASCAL-$5^i$ and +10.81 mIoU in COCO-$20^i$). 
On the other hand, L-ZSSS approaches show poor generalization.
Even though L-ZSSS approaches improved on the training domain (PASCAL), they performed slightly better (ZSVM, +1.12 mIoU) or worse (SPNet, -6.02 mIoU) than their backbone in the unseen domain (COCO). 

\textbf{Qualitative results}
\label{mscoco:qual}
Examples of MSCOCO 2017 in Figure \ref{figure:qual} contain three challenging cases of ZSSS:
multiple objects in different scales (the fourth and the fifth column);
a complex object in a noisy background (the sixth column);
objects to compare the generalization ability of models (the seventh and the eighth column).

\emph{Multi-scale Understanding}
Shown in the fourth column (vase) and the fifth column (motorcycle) in Figure \ref{figure:qual}, only SM-VCENet accurately recognizes multi-scale multiple objects in a scene.
Contrarily, DeepLab and L-ZSSS models (SPNet and ZSVM) fail to segment more than one object in a single image (vase).
In addition, they performs poorly in the case of multi-scaled objects (motocycle).
It is notable that L-ZSSS models produces worse prediction than their backbone (DeepLab).

\emph{Spatial Understanding}
The sixth column shows a case of a truck in a noisy background with less brightness and less contrast.
Understanding the complex scene requires detailed global context information.    
Our SM-VCENet accurately perceives the shape of the truck.
Oppositely, SPNet and ZSVM fail to recognize the truck although their backbone (DeepLab) finds it.

\emph{Generalization}
The seventh and the eighth column in Figure~\ref{figure:qual} carry examples of 'person' and 'bear' categories.
The training class set (PASCAL-$5^i$) contains 'person' and does not include 'bear'. 
SM-VCENet achieves qualitative performance improvements in the seen class (person) and the unseen class (bear) in the target domain. 
Although SPNet and ZSVM achieve significantly accurate performances in the trained domain (PASCAL VOC 2012), they fail to recognize the seen class (person) neither the unseen class (bear) in the target domain.
The performance degradation of SPNet and ZSVM means L-ZSSS fails to generalize to unseen domain.

\begin{figure}[t]
  \centering
    \includegraphics[width=0.40\textwidth]{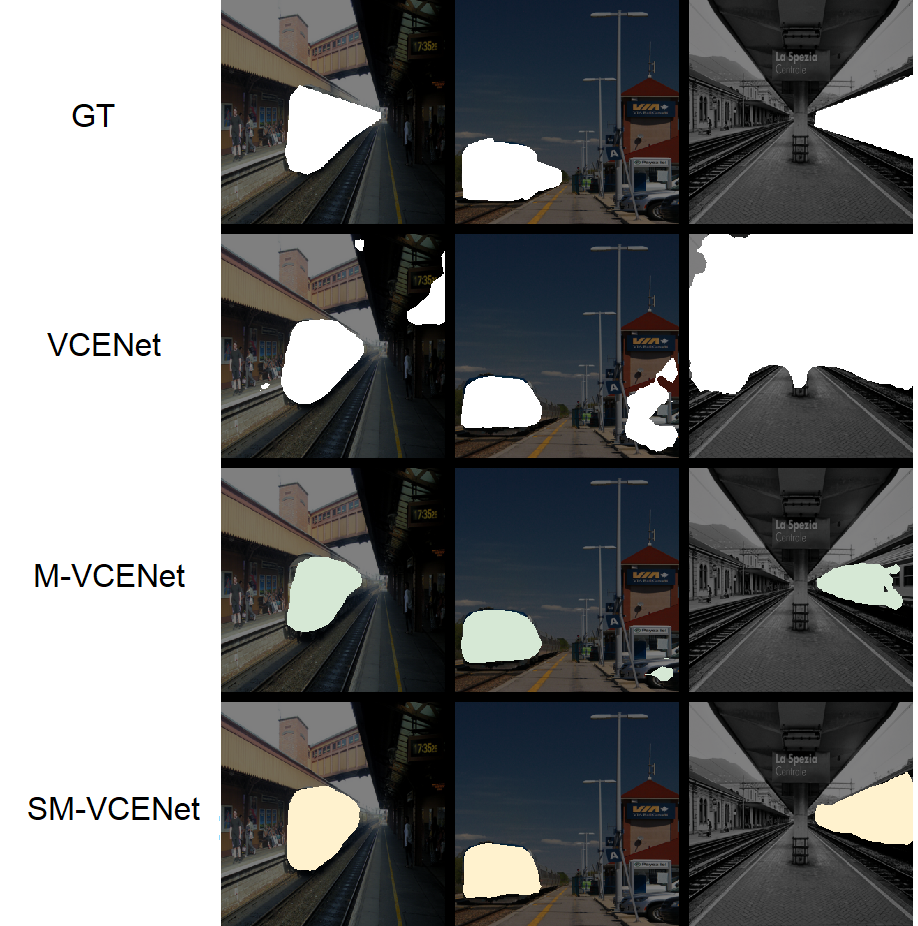} {}
  \caption{
Ablation study on MAM and SAM.
GT denotes ground truth.
VCENet is a compositional architecture of DeepLab and visual class branch.
M-VCENet and SM-VCENet are VCENet with MAM and both SAM and MAM.
}
\label{figure:ablation}
\end{figure}
\begin{table}[t]
\centering
\resizebox{.45\textwidth}{!}{
\begin{tabular}{ccccc}
\toprule
                 & M & S & P-mIoU & C-mIoU \\
\midrule
Deeplab          &   &   & 38.59  & 30.53  \\
VCENet           &   &   & 45.01 (+6.42)  & 34.43 (+3.90)  \\
M-VCENet         & \checkmark &   & 45.77 (+7.18)  & 40.81 (+10.28)  \\
SM-VCENet        & \checkmark & \checkmark & 46.51 (+7.92)  & 41.34 (+10.81)  \\
\bottomrule
\end{tabular}
}
\caption{
Ablation study of the proposed VCE, MAM and SAM.
P-mIoU and C-mIoU refer Pascal-$5^i$ mIoU score and COCO-$20^i$ mIoU score respectively.
}
\label{table:ablation}
\end{table}

\subsection{Ablation Study} 
\label{ablation}
We conducted ablation study of VCE, MAM and SAM. 
Table~\ref{table:ablation} shows accuracy improvements of VCE concept, multi-scale attention module (MAM) and spatial attention module (SAM). 
VCENet is a combined architecture of DeepLab and VCE branch that extracts class embedding with ImageNet knowledge.
The M and S columns of the table represent the presence of MAM and SAM, respectively.
In the both test sets (PASCAL-$5^i$ and COCO-$20^i$), VCE branch, MAM and SAM regularly increases the accuracy.
Figure~\ref{figure:ablation} shows the qualitative results on COCO-$20^i$ test images.
Compared with VCENet, MAM (M-VCENet) enhances understanding objects in various scales.
Furthermore, SAM (SM-VCENet) that grasps global context of the scene achieves clearer prediction than M-VCENet.

\section{Conclusion}
We presented a novel language-model-free zero-shot semantic segmentation framework (SM-VCENet), tackling limitations of L-ZSSS approaches. 
We first addressed that L-ZSSS approaches are vulnerable to domain shift as their class embedding is domain-agnostic.
To verify the problem, we presented that domain shift causes changes of visual class embedding while language-based class embedding is constant. 
In addition, our experiments in PASCAL2COCO including domain adaptation problem showed that L-ZSSS approaches have a limitation in generalization.
To overcome this limitation, we proposed language-model-free and query-domain-aware VCE in our SM-VCENet. 
Contrary to word-vector-driven class embedding in L-ZSSS, our VCE is vision-oriented and thus can take leverages of visual information.
Our SAM and MAM in SM-VCENet enhance the utilization of visual information achieving significant improvements in visually challenging scene understanding.
In the public benchmark PASCAL-$5^i$, our proposed SM-VCENet including VCE, SAM, and MAM records ZSSS SOTA. 
In our experiments in PASCAL2COCO, SM-VCENet presented generalization-robustness and superiority in visually challenging problems.
Our ablation study substantiated the effectiveness of VCE, SAM and MAM. 

\bibliography{egbib}

\end{document}